\documentclass[10pt,twocolumn,letterpaper]{article}

\usepackage{cvpr}
\usepackage{times}
\usepackage{epsfig}
\usepackage{graphicx}
\usepackage{amsmath}
\usepackage{amssymb}
\usepackage{enumitem}
\usepackage[ruled,linesnumbered]{algorithm2e}
\usepackage{setspace}

\usepackage[pagebackref=true,breaklinks=true,letterpaper=true,colorlinks,bookmarks=false]{hyperref}

 \cvprfinalcopy % *** Uncomment this line for the final submission

\def\cvprPaperID{1271} % *** Enter the CVPR Paper ID here

\ifcvprfinal\fi
\begin{document}

%%%%%%%%% TITLE
\title{Factors in Finetuning Deep Model for Object Detection with Long-tail Distribution}

\author{Wanli Ouyang, Xiaogang  Wang, \\
The Chinese University of Hong Kong\\
{\tt\small wlouyang, xgwang@ee.cuhk.edu.hk}
\and
Cong  Zhang,  Xiaokang  Yang\\
Shanghai Jiaotong University\\
{\tt\small zhangcong0929, xkyang@sjtu.edu.cn}
}

\maketitle
%\thispagestyle{empty}

%%%%%%%%% ABSTRACT
\begin{abstract}
 Finetuning from a pretrained deep model is found to yield state-of-the-art performance for many vision tasks. This paper investigates many factors that influence the performance in finetuning for object detection. 
 There is a long-tailed distribution of sample numbers for classes in object detection. Our analysis and empirical results show that classes with more samples have higher impact on the feature learning. And it is better to make the sample number more uniform across classes. Generic object detection can be considered as multiple equally important tasks. Detection of each class is a task.
These classes/tasks have their individuality in discriminative visual appearance representation. Taking this individuality into account, we cluster objects into visually similar class groups and learn deep representations for these groups separately. A hierarchical feature learning  scheme is proposed. In this scheme, the knowledge from the group with large number of classes is transferred for learning features in its sub-groups. Finetuned on the GoogLeNet model, experimental results show 4.7\% absolute mAP improvement of our approach on the ImageNet object detection dataset without increasing much computational cost at the testing stage. 
\end{abstract}

%%%%%%%%% BODY TEXT
\section{Introduction}
Finetuning refers to the approach that initializes the model parameters for the target task from the parameters pretrained on another related task. Finetuning from the deep model pretrained on the large-scale ImageNet dataset  is found to yield state-of-the-art performance for many vision tasks such as tracking \cite{wang2015visual}, segmentation \cite{girshick2014rich}, object detection \cite{szegedy2014going, ouyang2015deepid, ouyang2015learning}, action recognition \cite{karpathy2014large}, and human pose estimation \cite{chu2015multi}. When finetuning the deep model for object detection, however, we have two observations.

\begin{figure}
\centering
%\vspace{-10pt}
\includegraphics[width=1\linewidth]{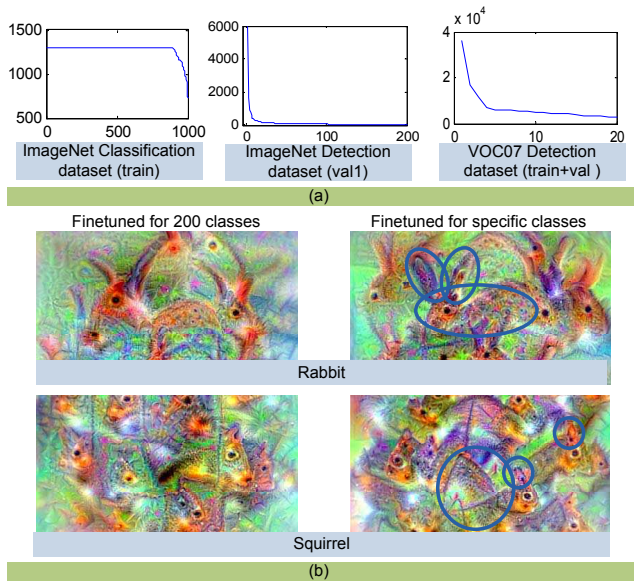}
%\centerline{\epsfig{figure=./eps/Visio-Fig1-3.eps,width=8cm}}
%\vspace{-10pt}
\caption{The number of samples in y-axis sorted in decreasing order for different classes in x-axis on different datasets  \textbf{(a)} and the models obtained using different strategy \textbf{(b)}. Long-tail property is observed for ImageNet and Pascal object detection dataset in (a). Models are visualized using the DeepDraw \cite{web:deepDraw}. Compared with  the model on the left in (b)  finetuned for all the 200 classes in ImageNet detection dataset, the model finetuned for specific classes on the right column in (b) is better in representing rabbit and squirrel. \textbf{Best viewed in color}.}
\label{Fig:SampleNumCls}
\end{figure}

The first is the long-tail property.
The ImageNet image classification dataset is a well compiled dataset, in which objects of different classes have similar number of samples. In real applications, however, we will experience the long-tail phenomena, where small number of object classes appear very often while most of the others appear more rarely. For segmentation, pixel regions for certain classes appear more often than the regions for other classes. For object detection, some object classes such as person have much more samples than the other object classes like sheep for both PASCAL VOC \cite{Everingham:PacalVOC} and ImageNet \cite{ILSVRC15} object detection dataset, as shown in Fig. \ref{Fig:SampleNumCls}(a).  More examples and discussion on the long-tail property is given in a recent talk given by Bengio \cite{Bengio:Battle}. For detection approaches using hand-crafted features \cite{LatSVMObj, vandeSandeCVPR2014}, feature extraction is separated from the classifier learning task. Therefore, the feature extraction is not influenced by the long-tail property. For deeply learned features, however, the feature learning will be dominated by the object classes with large number of samples so that the features are not good for object classes with fewer samples in the long tail. We analyze the influence of long tail in learning the deep model using the ImageNet object detection dataset as a study case. We find that even if around 40\% positive samples are left out from this dataset for feature learning, the detection accuracy is improved a bit if the number of samples among different classes is more uniform.

The second is in learning specific feature representations for specific classes. The detection of multiple object classes  is composed of multiple tasks. Detection of each class is a task. At the testing stage, detection scores of different object classes are independent. And evaluation of the results are also independent for these object classes. Existing deep learning methods consider all classes/tasks jointly and learn a single feature representation \cite{girshick2014rich, szegedy2014going}. Is this shared representation the best for all object classes? Objects of different classes have their own discriminative visual appearance. If the learned representation can focus on specific classes, e.g. mammals, the learned representation is better in describing these specific classes. For example, the model finetuned for 200 classes only focuses on the head of  rabbit and squirrel, as shown in Fig. \ref{Fig:SampleNumCls} (b).  In comparison, if the model can focus on these mammals, the model can also learn representation for the body and ear shape of rabbit, and the tail and ear shape of squirrel, as highlighted by the blue ellipses in Fig. \ref{Fig:SampleNumCls} (b). In this paper, we propose a scheme that inherits the shared representation and learns the specific representation for the specific subset of classes, e.g. mammals.

The contribution of this paper is as follows:

1. Analysis and experimental investigation on the factors that influence the effectiveness of finetuning. The investigated factors include the influence of the pretraining and finetuning on different layers of the deep model, the influence of the long tail, the influence of the training sample number, the effectiveness of different subsets of object classes, and the influence from the subset of training data.

 2. A cascaded hierarchical feature learning approach. In this approach, object classes are grouped. Different models are used for detecting object classes in different groups. The model gradually focuses on the specific group of object classes. The knowledge from the larger number of generic classes is transferred to the smaller number of specific classes through hierarchical finetuning. The cascade of the models saves the computational time and helps the models to focus on hard samples. Through cascade, each model only focuses on around 6 candidate regions per image. 
 With the proposed feature learning approach, 4.7\% absolute  mAP increase is achieved on the ImageNet object detection dataset.

\section{Related work}
The long-tail property is noticed by researchers working on scene parsing \cite{yang2014context} and zero-shot learning \cite{norouzi2013zero}. Yang \emph{et~al.} \cite{yang2014context} expand the samples of rare classes and  achieve more balanced superpixel classification results. Norouzi \emph{et~al.}  \cite{norouzi2013zero}  use the semantically similar object classes to predict the unseen classes of images. Deep learning is considered as a good representation sharing approach in the battle against the long tail \cite{Bengio:Battle}. The influence of long tail in deep learning, to our knowledge, is not investigated. We provide analysis and experimental investigation on the influence of the long tail in learning features. Our investigation provides knowledge for training data preparation in deep learning.

Deep learning is found to be effective in many vision tasks \cite{wang2015visual, chu2016structure, MultiGPUCaffe2015, wang2015STCT, Ouyang2014MultiSource, Ouyang2013JointDeep, Ouyang:DBNHuman, zhu2013deep, Luo:DeepFace, sun2014deep, Sun2013Deep, dong2014learning, zhang2014facial,shao2015deeply}.

Deep learning is applied for object detection in many works \cite{girshick2014rich, sermanet2013overfeat, lin2013network, he2014spatial, szegedy2014going, zhang2015improving, yan2015object, girshick2015fast, ren2015faster, lenc2015r, redmon2015you, ouyang2015deepid, ouyang2016learning, Zeng2013Multi, Kang2016Object}. Existing works mainly focus on developing new deep models \cite{sermanet2013overfeat, lin2013network, szegedy2014going, girshick2015deformable, redmon2015you} and better object detection pipeline \cite{girshick2015fast, ouyang2015deepid, yan2015object, ren2015faster, lenc2015r, redmon2015you}. These works use one feature representation for all object classes. When using the hand-crafted features, the same feature extraction mechanism is used for all object classes \cite{Azizpour:DetStrong, yanfastest, LatSVMObj}. In our work, however, different object classes use different deep models so that the discriminative representations are specifically learned for these object classes.

Our work is also different from the model ensemble used in \cite{szegedy2014going, yan2015object, ouyang2015deepid}. In model ensemble, the detection score for an object class is from multiple deep models with different parameters or different network architectures. The detection score for an object class is from only one model in our approach. Therefore, our approach is complementary to model ensemble in further improving accuracy.

Cascade is used in many object detection works \cite{Felzenszwalb:cascade10, Dollar:Crosstalk, Viola:HaarConf}. We use cascade to speed-up the testing stage.

\section{Factors in finetuning for ImageNet object detection}

\subsection{Baseline model}
\label{sec:Baseline}
\label{Sec:Dataset}
\emph{Region proposal.}
The learned detector is used for classifying each candidate region as containing certain object class or not. In this paper we use the selective search \cite{Smeulders:SelectiveSearch} for obtaining candidate regions. By default, we use the bounding box rejection approach in \cite{ouyang2015deepid} so that around 6\% candidate regions from selective search are retained. 

\emph{Training and Testing data.} 
We use the large scale ImageNet detection dataset for training and testing. 
The ImageNet ILSVRC2013 detection dataset is split into three
sets: train13 (395,918), val (20,121), and test (40,152), where
the number of images in each set is in parentheses. Based on the ILSVRC2013 dataset, the extra train14 (60,658), is collected in the ILSVRC2014 dataset. There are 200 object classes for detection in this dataset.
%The train14, val and test are fully annotated, meaning that in each image all instances from all 200 classes are labeled with bounding boxes. In contrast, the train13 images are not fully annotated, i.e. train images may contain objects that are not annotated. 
val and test splits are drawn from the same image distribution. To use val for both training and validation, val is split into “val1” and “val2” in \cite{girshick2014rich}. The split is copied from \cite{girshick2014rich} for our experiments. The test set is not available for extensive evaluation. Therefore, we have to resort to the val2 data, which contains around 10,000 images. If not specified, val2 images are used for evaluation, and the train13, val1, and train14 images are used for training. If not specified, we use the selective search to obtain negative and extra positive bounding boxes in val1 and the ground truth positive bounding boxes in train13 and train14.

\emph{Network, pretraining, and fintuning.} 
The GoogLeNet \cite{szegedy2014going} is shown to be the state-of-the-art in many recent works \cite{ouyang2015deepid, szegedy2014going, yan2015object} for ImageNet object detection. We use exactly the same model structure as that in \cite{szegedy2014going}. 
%GoogLeNet is pretrained on the 1000-class ImageNet classification task using the bounding box annotations and then finetuned for the object detection task, as in \cite{ouyang2015deepid}. As reported in \cite{ouyang2015deepid}, for the same model structure,  the model pretrained with bounding box annotations performs better than the model pretrained without bounding box annotations.  % If the pretrained model is directly used for extracting features without fine-tuning, the mAP is 33\%.  
The pretrained GoogLeNet with bounding box annotations provided online \footnote{\url{www.ee.cuhk.edu.hk/~wlouyang/projects/imagenetDeepId/index.html}} is used for finetuning. The mAP on val2 is 40.3\% in our four of five trials, another trial has 40.4\% mAP.  Therefore, the pretrained model we use with 40.3\% mAP after finetuning is better than that in \cite{szegedy2014going}, which is 38.8\%.  At the finetuning stage, aside from replacing the CNN’s 1000-way pretrained classification layer with a randomly initialized (200 + 1)-way softmax classification layer (plus 1 for background), the CNN architecture is unchanged.

%The val and test splits are drawn from the same image distribution. These images are scene-like and similar in complexity (number of objects, amount of clutter, pose variability, etc.) to PASCAL VOC images. The val and test splits are exhaustively annotated, meaning that in each image all instances
%from all 200 classes are labeled with bounding boxes.

\emph{SVM learning.}
After the features are learned, one-vs-rest linear SVMs are learned for obtaining the detectors for each object class,  the same as that in \cite{girshick2014rich}.
%The positive data for SVM training is from train13, val1, and train14 ground-truth bounding box, the negative data  for SVM training is from 5000 images of val1,
Since this paper focuses on learning deep model, training data preparation for SVM is kept unchanged  for all experiments, although we will investigate different training data preparation for deep model learning.

\emph{Summary.} Pretrained with bounding box annotations, the baseline GoogLeNet has 40.3\% mAP on val2 when trained using ILSVRC14 detection data with selective search for region proposal. This deep model is finetuned by 200+1 softmax loss and then linear SVM is used for learning the classifier based on the learned deep  model.

For the experiments we conduct in this paper, we only change one of the factors while keeping others the same as the baseline.

\subsection{Investigation on freezing the pretrained layer in finetuning}
In this experiment, we investigate how finetuning specific layers influences the detection performance. Given the pretrained GoogLeNet, we freeze the parameter of certain layers and only finetune parameters of the remaining layers. The experimental results are shown in Table \ref{Tab:layers}.
There are 11 modules in the GoogLeNet: two convolutional layers, i.e. conv1 and conv2, and nine inception modules, i.e. icp (3a), icp (3b), icp (4a), icp (4b), icp (4c), icp (4d), icp (4e), icp (5a), and icp (5b).  If we freeze all the 11 modules and use the features learned from the pretrained model for learning SVM classification, the mAP is 33.0\%, much worse than finetuning of all 11 modules that has mAP 40.3\%. Finetuning all the 11 modules has the same mAP as  freezing the conv1-icp(4a) during finetuning. These frozen modules are extracting general low level features, which have been well-learned by the pretrained model. Therefore, it is so not necessary to finetune these modules. The mAP only drops by 0.7\% even if we fix the eight modules conv1-icp(4d), which takes 43\% the number of parameters, and 80\% the number of operations. As we freeze more and more modules to higher levels, the mAP decreases more and more rapidly. The upper layers are more responsible for discriminating semantic objects. Therefore, finetuning of the upper layers have more impact on the detection performance.

\begin{table*}[]
\centering
{\small
\begin{tabular}{c|cccccc}
\hline
Num. modules frozen & 0 & 3             & 6             & 7             & 8             & 9             \\
Modules frozen          &none &conv1-icp(4a) & conv1-icp(4d) & conv1-icp(4e) & conv1-icp(5a) & conv1-icp(5b) \\
mAP                    & 40.3& 40.3        & 39.6        & 38.8        & 36.5        & 33
\end{tabular}
}
\caption{Detection mAP ($\%$) on val2 when freezing modules in GoogLeNet.}
\label{Tab:layers}
\end{table*}

\subsection{Investigation on training data preparation}
In this section, we investigate the use of different training data for learning features. The same as the baseline setting in Section \ref{sec:Baseline}, train13, train14 and val1 are used for learning the SVM. In this way, only the learned features are the factors in influencing the detection performance.

\subsubsection{Investigation on different subset of training data}
As illustrated in Section \ref{Sec:Dataset}, there are three different subsets of training data. The performance of single subset and leave-one-subset-out is shown in Table \ref{Table:DatasetChoice}. 
Experimental results show that train13 is not so effective in learning features when compared with train14 and val1. The val1, val2 and test images are scene-like. The train13 images are drawn from the ILSVRC2013 classification image distribution. It has a skew towards images of a single centered object. The mismatch in train13 and val leads to the lower mAP in using train13 only.

If positive samples are from train14, the model trained using negative samples from val1 has mAP 35.2\%, while the model using the negative samples from train14 has mAP 39.6\%. Therefore, for the same positive samples, it is better to use negative samples from the same image instead of from other images for learning the model.

If the positive samples and the negative samples are from the same images, val1 has mAP 39\% and train14 has mAP 39.6\%. There are 60,658 train14 images  and 9,887 val1 images. The increase of training images by about 6 times only results in 0.6\% mAP improvement. We find that train14, although claimed to be fully annotated, still has many objects not annotated. The unannotated objects are much less on val1 and val2. The noise in having potential objects not annotated is one of the reasons for the small increase in mAP with the large increase in training images. 
We will further investigate the relationship between the number of samples and mAP in Section \ref{Sec:LongTail}.

\begin{table*}
\setlength{\tabcolsep}{2pt}
\centering
{\small
\begin{tabular}{c|ccccccc}
\hline positive & train13 & val1(s) & train14(g) & train14(s) & train14(g)+val1(s) & train13+train14(g) & train13+val1(s) \\
negative & val1    & val1     & val1       & train14    & val1               & val1               & val1            \\
\hline mAP      & 37.5    & 39       & 35.2       & 39.6       & 39.3               & 37.2               & 40.1           

\end{tabular}
}
\caption{Detection mAP ($\%$) on val2 trained from different combination of training data. The performance of using train13+val1+train14 is 40.3\%. (s) denote the augmentation of positive data by boxes from selective search. (g) denotes the use of only ground-truth data. }
\label{Table:DatasetChoice}
\vspace{-10pt}
\end{table*}

\subsubsection{The long-tail property}
\label{Sec:LongtailAna}
\begin{figure}
\centering
%\vspace{-10pt}
\includegraphics[width=0.9\linewidth]{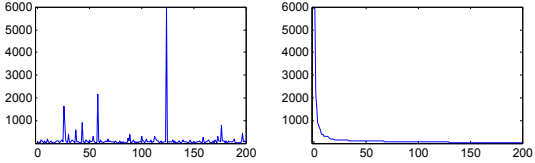}
%\centerline{\epsfig{figure=./eps/Visio-Fig1-3.eps,width=8cm}}
%\vspace{-10pt}
\caption{The number of annotated positive samples in val1 as a function of the object class index in ImageNet (left) and the number of samples for each class sorted in decreasing order (right). The three classes largest in sample number are person (6,007), dog (2,142) and bird (1643), where the number of samples for each class is in parentheses. In comparison, the three classes smallest in sample number are hamster (16), lion (19), and centipede (19).}
\label{Fig:SampleNumCls2}
\end{figure}

Fig. \ref{Fig:SampleNumCls2} shows the number of samples in val1 for the 200 object classes. It can be seen from Fig. \ref{Fig:SampleNumCls2} that  the number of samples varies a lot for different classes. When the object classes are sorted by the number of samples, we observe the long-tail property. 59.5\% ground-truth samples are from 20 object classes with largest sample number. Similar statistics are observed in the val2 data. Although we are not provided with the test data annotations, it is reasonable to assume that the test data also has similar long-tail property.
 In order to make the number of samples more uniformly distributed, the number of samples from train13 is constrained to be less than or equal to  1,000 in \cite{girshick2014rich}. With this constraint, 49.5\% ground-truth samples are from 40 object classes with largest sample number when considering the train13, val1 and train14 data altogether. The long tail still exists.

The softmax (cross entropy) loss used for learning the deep model is as follows:
%\vspace{-5pt}
\begin{equation}
\begin{split}
L &= -\sum_{n=1}^{N}{\sum_{c=1}^{C}{t_{n,c} \log p_{n,c}}}, \\
 \textrm{where } p_{n,c} &= \frac{e^{net_{n,c}}}{e^ {\sum_{c=1}^{C}{net_{n,c}}}}.
%&p_{n,c} = \frac{\exp{(\mathbf{w}_c^\textrm{T} \mathbf{h_i}})}{\sum_c{\exp{(\mathbf{w}_c^\textrm{T} \mathbf{h_i})}}}
\end{split}
\end{equation}
 $t_{n,c}$ denotes the target label and  $p_{n,c}$ denotes the prediction for the $n$th sample and $c$th class. $t_{n,c}=1$ if the $n$th sample belongs to the $c$th class, $t_{n,c}=0$ otherwise. $net_{n,c}$ is the classification prediction from the neural network. Denote $\theta$ as the parameters to be learned from the network, the derivative is as follows:
{\small
%\vspace{-5pt}
\begin{equation}
\begin{split}
&\frac{\partial L}{\partial \theta} =\sum_{n,c} ( p_{n,c} - t_{n,c}) \cdot  \frac{\partial net_{n,c} }{\partial \theta}. \\
\end{split}
\label{eq:SoftMaxGrad}
\end{equation}
}
It can seen from (\ref{eq:SoftMaxGrad}) that the gradient of the parameters is influenced by two factors. First,  the accuracy of $p_{n,c}$ in predicting $t_{n,c}$. The more accurate $p_{n,c}$ is, the smaller the gradient in back-propagation (BP) for the $n$th sample. Second, the number of samples belonging to class $c$. Suppose the prediction error  $( p_{n,c} - t_{n,c})$ in (\ref{eq:SoftMaxGrad}) is of similar magnitude for all samples. If the class bird has 16,00 samples but the class hamster  has only 16 samples, then the magnitude of the gradient from bird will be around 100 times of  the magnitude of the gradient from hamster. In this case, although the network representation is shared by different classes, the network parameters will be dominated by the class bird which has much more samples. This is fine for applications where the importances of classes are determined by their sample number. For applications like object detection, however, each class is equally important. The features learned from deep model dominated by the class bird may not be good for the class hamster.

\subsubsection{Experimental results on the long-tail property}
\label{Sec:LongTail}
 In this experiment, we use train13, train14 and val1 as the training data, which are supposed to have $N_+$ positive samples/bounding-boxes and $N_-$ negative samples/bounding-boxes. We obtain subset from these data by the following three schemes:
\begin{enumerate}[leftmargin=12pt,noitemsep,nolistsep]
\item \emph{Rand-pos}. In this scheme, the $N_+$ positive boxes are reduced  to be $N_+'$ boxes by random sampling so that $N_+'/N_+ =r= \{2^{-1},2^{-2}, 2^{-3}, \ldots \}$. $r$ corresponds to the ratio of the remaining positive boxes. The negative boxes are kept unchanged.
\item \emph{Rand-all}. In this scheme, the numbers of positive and negative boxes are reduced to be $N_+'$ and $N_-'$ respectively  by random sampling so that $N_+'/N_+ =N_-'/N_- =r$. 
\item \emph{Pseudo-uniform}. In this scheme, the classes with samples larger than $N_{max}$ will be randomly sampled to have $N_{max}$ remaining samples. Classes with samples smaller than $N_{max}$ are untouched. We also require that the remaining samples divided by $N_+$ is $r$. Denote the number of positive boxes for class $c$ by $N_{+, c}$. After sampling, we have $N_{+, c}'$ positive boxes for class $c$. In this scheme, we have $r=(\sum_c{N_{+, c}'})/N_+, N_{+, c}' <= N_{max}$. 
\end{enumerate}
In the pseudo-uniform scheme, the number of positives samples for different classes becomes more uniform when $N_{max}$ is smaller. In the Rand-pos and Rand-all scheme, the long-tail property is preserved. 

\begin{figure}
\centering
%\vspace{-10pt}
\includegraphics[width=0.7\linewidth]{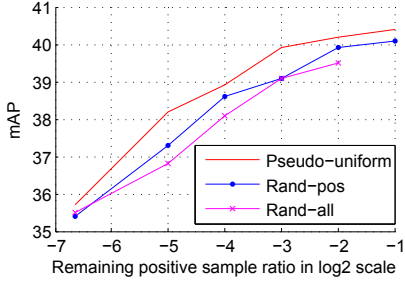}
%\centerline{\epsfig{figure=./eps/Visio-Fig1-3.eps,width=8cm}}
%\vspace{-10pt}
\caption{mAP on val2 as a function of the ratio $r$ for remaining positive boxes  for three different schemes -- rand-pos, rand-all, and pseudo-uniform. }
\label{Fig:SampleNum}
\end{figure}

 Fig. \ref{Fig:SampleNum} shows the experimental results on the three different schemes. Using all negative samples, we can see that pseudo-uniform performs better than rand-pos if the same number of positive samples are used. In fact, when $\log_2{r}=-1, -2, -3, -4, -5$, pseudo-uniform requires only half the number of positive samples to achieve the same mAP as rand-pos. For example, pseudo-uniform has mAP 39.9\% when $\log_2{r}=-3 (r = 12.5\%)$, and rand-pos has mAP 39.9\% when $\log_2{r}=-2 (r=25\%)$. The baseline that uses all samples has mAP 40.3\%. In the pseudo-uniform scheme, we observe small improvement (mAP 40.5\%) when $N_{max}=3,000$, in  which around 40\% positive samples are not used in the finetuning. If we keep all the training samples and enforce that the mini-batches in the stochastic gradient descent based BP should have uniform distribution in positive sample number, the mAP is 40.7\%. The approaches in increasing mAP are not used in the other experimental results of this paper for fair comparison.
 Therefore, our empirical results show that it is better to have uniform number of samples per class than long-tailed samples for learning features. 
 
 If both positive and negative boxes are randomly sampled using the scheme rand-all, the performance drops by 0.4\%-0.5\% compared with the rand-pos scheme that only samples positive boxes.

When $r=0.01$, rand-all have mAP 35.5\%. In this case, only around 34 positive boxes per class are used for finetuning. Finetuning (mAP 35.5\%) still has observable increase in mAP (2.5\%) compared with the model without fine-tuning (mAP 33\%). 

%It can also be observed from the curve in Fig. \ref{Fig:SampleNum} is that the mAP does not increase in log-linear scale with mAP. The increase in mAP is getting lower as the number of samples increases.

\subsection{Experimental study on subsets of object classes}
\label{Sec:SubsetClass}
There are 200 object classes in the ImageNet object detection challenge. In this section, we investigate using features learned from a subset of object classes for the other object classes. 
As analyzed in Section \ref{Sec:LongtailAna}, both the number of samples in each class and the estimation accuracy determine the gradient in learning the deep model. We conduct experiments on using subsets of object classes that have largest/smallest number of samples and largest/smallest accuracy. The results are shown in Table \ref{table:50100cls}.

%In this experiment, we investigate the use of subsets of object classes for the whole set of object classes. 
We can use the $C=\{50, 100, 150\}$ object classes having the largest accuracy for finetuning, and then use this model for extracting features for all the 200 classes and learning their SVMs.  In this way, the mAP is 37.9\% when $C=50$. Much better than the model without finetuning, which has mAP 33\%. Therefore, finetuning on the 50 classes has learned representations that can be shared by the other 150 classes that are not used for finetuning the deep model. For example, a learned feature that is good at describing the dog is good for describing the tiger. 

If we leave the 50 object classes with fewest samples out and use 150 object classes with most samples for finetuning, the mAP is 40.1\%. If all the 200 classes are used for finetuning, the mAP is 40.3\%.  The inclusion of the 50 object classes with the fewest samples in finetuning only increases mAP by  0.2\%. 
%Therefore, the 150 object classes with most samples dominate the parameter learning in the  finetuning stage. Similarly for the 150 classes with largest accuracy.

It can be seen from the results that the number of object classes used for finetuning is the key factor in influencing the mAP.  For example, the use of 50 object classes have only at most 37.9\% mAP in Table \ref{table:50100cls}. Among the 200-class positive boxes used for finetuning, these 50-class boxes have around 50\%  samples. In comparison, if the 50\% positive samples are randomly sampled from 200 classes for finetuning, the mAP is 40.1\%, as shown in Fig \ref{Fig:SampleNum}. In fact, even if only $6.25\%$ positive boxes are randomly sampled, the mAP is 38.6\% and performs better than the use of only 50 classes.
Among the four choices of subsets in Table \ref{table:50100cls}, the choice of the $C$ least accurate object classes has the lowest mAP. Thus it is the worst choice in obtaining features that can be shared by other object classes. 

This section investigates the use of $C$ classes for 200 classes in finetuning. The next section investigates the use of $C$ classes for $C$ classes in finetuning.

%In this experiment, we investigate the use of a subset of object classes for this subset of object classes.  We can use the $C=\{50, 100, 150\}$ object classes having the largest accuracy for finetuning based on the 1000-class pretrained mode, and then use this model only for these $C$ classes and learning their SVMs. The \emph{$\Delta$mAP for specific class}  in Table \ref{table:50100cls} denotes the improvement in absolute mAP for these $C$ classes when compared with the basesline that finetunes the GoogleNet for 200 classes. In this way, we find that the mAP for the specific $C$ classes are consistently improved. 
%Therefore,  using the model  finetuned specifically for the $C$ classes is better than using the model the finetuned for 200 classes.
%We find that the classes with smaller accuracy also have small sample number compared with classes with larger number. For example, the 50 classes with lowest accuracy have only 13.5\% samples while the 50 classes with highest accuracy have 47\% samples. The model the finetuned for 200 classes have skew towards representations for the classes with larger sample number. 
%This is the season why the classes with smaller numbers get higher improvement in mAP when specifically learning representations for them.

\begin{table*}[]
\setlength{\tabcolsep}{2pt}
\centering
{\small
\begin{tabular}{c|ccc|ccc|ccc|ccc}
\hline       Choice                                                         & \multicolumn{3}{c}{Largest accuracy} & \multicolumn{3}{c}{Smallest accuracy} & \multicolumn{3}{c}{Largest number} & \multicolumn{3}{c}{Smallest number} \\
\hline
Num. cls             & 150        & 100        & 50         & 150         & 100        & 50         & 150        & 100       & 50        & 150        & 100        & 50        \\
pos num ratio & 86.50\%    & 65.80\%    & 47.00\%    & 53.00\%     & 34.20\%    & 13.50\%    & 91.20\%    & 79.59\%   & 62.40\%   & 37.61\%    & 20.42\%    & 8.80\%    \\
mAP      & 40.1       & 39.4       & 37.9       & 39.6        & 38.3       & 35.9       & 40.1       & 39.1      & 37.9      & 39.7       & 39.2       & 37.2      \\
%$\Delta$mAP for specific class   & 0.29\%     & 0.36\%     & 0.82\%     & 0.27\%      & 0.60\%     & 1.30\%     & 0.25\%     & 0.05\%    & 0.14\%    & 0.14\%     & 1.07\%     & 1.04\%   
\end{tabular}
}
\caption{Object detection accuracy in mAP when finetuned using different number of classes. \emph{Num. cls} denotes the number of classes used for finetuning. \emph{pos num ratio} denotes ratio, \ie the number of positive samples for class subset choice divided by the number of the all positive samples. \emph{Largest/least accuracy} denotes the use of the most/least accurate 50/100/150 classes for finetuning. Softmax accuracy of the training data is used for evaluating accuracy.  \emph{Largest/least number} denotes the use of the  50/100/150 classes with the largest/smallest training sample number for finetuning.}
\label{table:50100cls}
\vspace{-10pt}
\end{table*}

\section{Cascaded hierarchical feature learning for object detection}
\subsection{Grouping objects into hierarchical clusters}
\begin{figure}
\centering
%\vspace{-10pt}
\includegraphics[width=0.9\linewidth]{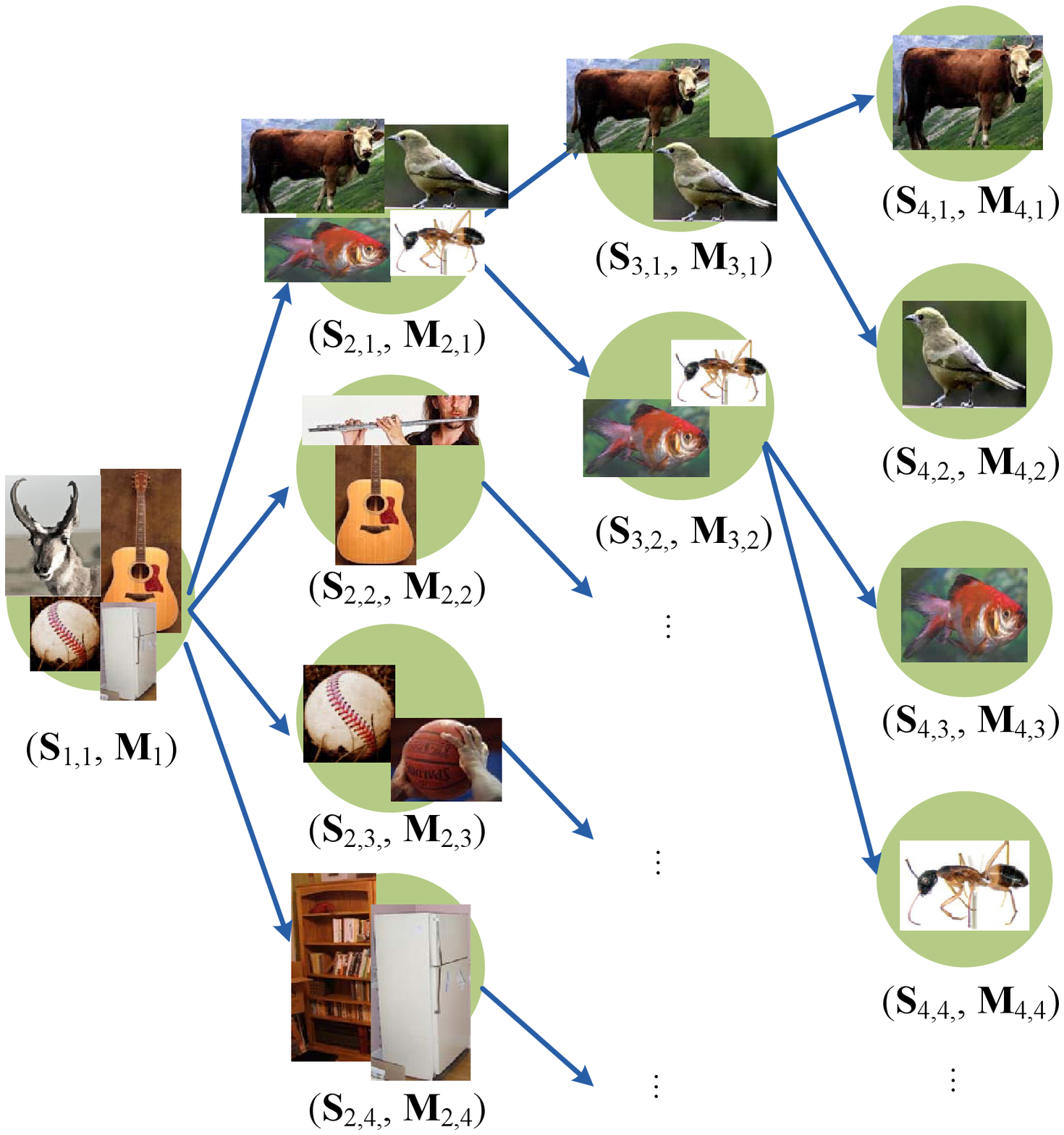}
%\centerline{\epsfig{figure=./eps/Visio-Fig1-3.eps,width=8cm}}
%\vspace{-10pt}
\caption{Grouping object classes into hierarchical clusters $\mathbb{S}_{l, j_l}$ and finetuning them to obtain multiple models $\mathbf{M}_{l, j_l}$. }
\vspace{-10pt}
\label{Fig:Cluster}
\end{figure}

The 200 object classes are grouped into hierarchical clusters. Our approach is not constrained to any clustering method. In Section \ref{Sec:ExpClusterMethod}, we will investigate different clustering methods, in which we find visual similarity to be the best in detection accuracy. Thus we use visual similarity as the example for illustration. The visual similarity between classes $a$ and $b$ is as follows:
\begin{equation}
\begin{split}
Sim(a,b) = \sum_{i=1}^{N_i} \sum_{j=1}^{N_j} <\mathbf{h}_{a,i}, \mathbf{h}_{b,j}>/N_iN_j , \\
\end{split}
\end{equation}
where $\mathbf{h}_{a,i}$ is the last GoogleNet hidden layer for the $i$th training sample of class $a$, $\mathbf{h}_{b,j}$ is for the  $j$th training sample of class $b$. $<\mathbf{h}_{a,i}, \mathbf{h}_{b,j}>$ denotes the inner product between $\mathbf{h}_{a,i}$ and $\mathbf{h}_{b,j}$. With the similarity between two classes defined, we use the approach in \cite{zhang2012graph} for grouping object classes into hierarchical clusters. At the hierarchical level $l$, denote the $j_l$th group by $\mathbb{S}_{l, j_l}$. In our implementation, $l = {1, \ldots, L}$, $L=4$, $j_l = \{1, \ldots, J_l\}$, $J_1=1, J_2=4, J_3 = 7, J_4 = 18$. Since there are 200 object classes in ILSVRC object detection, initially, $\mathbb{S}_{1, 1} = \{1, \ldots, 200\}$.  On average, there are 200 object classes per group at level 1, 50 classes per group at level 2, 29 classes per group at level 3, and 11 classes per group at level 4.
The hierarchical cluster result is shown in Fig. \ref{Fig:Cluster} by a few exemplar classes. In Fig. \ref{Fig:Cluster}, we have $\mathbb{S}_{1}=\mathbb{S}_{2,1}\cup \mathbb{S}_{2,2}\cup \mathbb{S}_{2,3}\cup \mathbb{S}_{2,4}$ and $\mathbb{S}_{2,1}=\mathbb{S}_{3,1}\cup \mathbb{S}_{3,2}$. In the hierarchical clustering results, the parent node $par(l, j_l)$  and children set ch$(l, j_l)$ of a node $(l, j_l)$ are defined such that  $\mathbb{S}_{l+1, j'} \subset  \mathbb{S}_{l, j_l},$ $\forall (l+1, j') \in$ ch$(l,j_l)$, $\mathbb{S}_{l, j_l}=\cup_{(l+1, j') \in \textrm{ch}(l, j_l)} \mathbb{S}_{l+1, j'}$, and $ \mathbb{S}_{l, j_l} \subset  \mathbb{S}_{l-1, par(l,j_l)}$. Therefore, a hierarchical tree structure is defined as shown by examples in Fig.  \ref{Fig:Cluster}.

\subsection{Our approach at the Testing Stage}
Our approach at the testing stage is described in Algorithm \ref{alg:testing}. 
In this approach, a testing sample is evaluated from root to leaves on the tree. At the node $(l, j_l)$, the detection scores for the classes in group $\mathbb{S}_{l,j_l}$ are evaluated (line \ref{alg:testscore} in Agorithm \ref{alg:testing}). 
These detection scores are used for deciding if the children nodes ch$(l, j_l)$ need to be evaluated (line \ref{alg:testcas} in Agorithm \ref{alg:testing}).
For the child node $(l+1,j') \in$ ch$(l, j_l)$,  if the maximum detection score among the classes in $\mathbb{S}_{l+1,j'}$ is smaller than a threshold $T_l$, this sample is not considered as a positive sample in class group $\mathbb{S}_{l+1,j'}$,  and then the node $(l+1,j')$ and its children nodes are not evaluated.  $T_l$ chosen so that the recall on val1 is not influenced much and a large number of candidates can be rejected.
 For example, initially the detection scores for 200-classes $\{y_{c}\}_{c \in \mathbb{S}_{1, 1}}$ are obtained at the node $(1,1)$ for a given sample of class bird. These 200-class scores are used for accepting this sample as an animal $\mathbb{S}_{2,1}$ and rejecting this sample as ball $\mathbb{S}_{2,2}$, instrument $\mathbb{S}_{2,3}$ or furniture $\mathbb{S}_{2,4}$. And then the scores $\{y_{c}\}_{c \in \mathbb{S}_{2, 1}}$  of animals are used for accepting the bird sample as vertebrate and rejecting it as invertebrate. Therefore, each node focuses on rejecting the sample as not belonging to a group of object classes. Finally, only the  groups that are not rejected have the SVM scores for their classes (line \ref{alg:testscore2} in Agorithm \ref{alg:testing}).

{
\setlength{\intextsep}{0pt}
\begin{algorithm}[h]
%\setstretch{1}
\KwIn{\{$ \mathbf{x}$, the testing sample.\\
~~~~~~~~~~~~ $\{\mathbb{S}_{l,j_l}\}$, hierarchical clusters of object classes. \\
~~~~~~~~~~~~ $\mathbf{M}_{l,j_l}$, the models.  \\
~~~~~~~~~~~~ \}}
 \KwOut{ \{$\mathbf{y}=[y_1, \ldots, y_C]$, the detection score of $\mathbf{x}$ \} }
$f_{1,1} =$ true \;
 $f_{l,j_l}=$ false for $l=2, \ldots, L$, $j_l = 1, \ldots, J_l$ \;
 \For{ $l$ = 1 to $L$ }{
\For{ $j_l$ = 1 to $J_l$ }{
\If{$f_{l,j_l}$ }{
  Get scores $\{y_{c}\}_{c \in \mathbb{S}_{l, j_l}}$ of $\mathbf{x}$ using $\mathbf{M}_{l, j_l}$ \label{alg:testscore} \;
%  $s=\max_{c \in \mathbb{S}_{l, j_l}}s_{c, l}$ \;
% If $s < T_l$, $f_{l+1,j'}=$ false $\forall j'\in$ ch$(j_l)$ \;
\For {$(l+1,j')\in$ ch$(j_l)$}{
 \textbf{If}  $\max_{c \in \mathbb{S}_{l+1, j'}}y_{c} \geq T_l$, \textbf{then}  $f_{l+1,j'}=$ true \label{alg:testcas} \;
 }
}
}\label{alg:endfor}
}
 $y_c =  s_{c, L}$ {for} $c \in \mathbb{S}_{L, j_{L}}$ if  $f_{L,j_{L}}$ is true;  $y_c = -\infty$ otherwise. \label{alg:testscore2} \\ 
  \caption{Our Approach at the Testing Stage. }
\label{alg:testing}
\end{algorithm}
}

\begin{algorithm}
\setstretch{1}
\KwIn{\{ $ \Psi = \{ \mathbf{x} \}$ training samples.\\
~~~~~~~~~~~~ $\{\mathbb{S}_{l,j_l}\}$, hierarchical clusters of object classes .\\
~~~~~~~~~~~~ $\mathbb{X}_{0,1,+}$, set of all positive samples.\\
~~~~~~~~~~~~ $\mathbb{X}_{0,1,-}$, set of all negative samples .\\
~~~~~~~~~~~~ $\mathbf{M}_{0,1}$, pretrained deep model.\\ 
~~~~~~~~~~~~\}}
\KwOut{\{$\mathbf{M}_{l,j_l}$, the finetuned models.\}~~~~~~~~~~~~ }
\For{ $l$ = 1 to $L$ }{
\For{ $j_l$ = 1 to $J_l$}{
 $\mathbf{M}_{l, j_l}= \mathbf{M}_{l-1, par(j_l)}$ \label{Alg:ini} \;
Finetune $\mathbf{M}_{l, j_l}$ using $\mathbb{X}_{l,j_l,+}$ and $\mathbb{X}_{l,j_l,-}$ \label{Alg:finetune}\;
\For{ $(l+1,j') \in ch(j_l)$}{
%~~~~~ ~~~~~~ ~~~~(choose positive and cascade negative) \nonumber \\ 
Use $\mathbf{M}_{l, j_l}$ to obtain detection scores  \{$y_{\mathbf{x}}\!=\!\max_{c \in \mathbb{S}_{l+1, j'}}y_{c}(\mathbf{x})|  \mathbf{x} \in \mathbb{X}_{l,j_{l},-}\}$ \;
$\mathbb{X}_{l+1,j',-}\!\!=\!\{\mathbf{x}| \mathbf{x}\in \mathbf{x}_{l,j_{l},-}\! \And \! y_{\mathbf{x}} > T_l\}$ \label{Alg:Pos} \;
 $\mathbb{X}_{l+1,j',+}\!\!=\!\{\mathbf{x}|\mathbf{x} \textrm{ is a class in } \mathbb{S}_{l+1,j'}\}$  \label{Alg:Neg}\;
}
}
}    
  \caption{Hierarchical Learning of the models. }
\label{alg:training}
\end{algorithm}

\subsection{Hierarchical Feature Learning}
The proposed feature learning approach is described in Algorithm \ref{alg:training}. 
Each node $(l, j_l)$ corresponds to a group of object classes $\mathbb{S}_{l, j_l}$. 
For the node $(l, j_l)$, a deep model $\mathbf{M}_{l, j_l}$ is finetuned using the model of its parent node $\mathbf{M}_{l-1, par(j_l)}$ as initial point (lines \ref{Alg:ini}-\ref{Alg:finetune} in algorithm \ref{alg:training}).
When finetuning the model  $\mathbf{M}_{l, j_l}$, the positive samples are constrained to have class labels in the group  $\mathbb{S}_{l, j_l}$ (line \ref{Alg:Pos} in Algorithm \ref{alg:training}), and the negative samples are constrained to be accepted by the its parent node (line \ref{Alg:Neg} in Algorithm \ref{alg:training}). 
Therefore, only a subset of object classes are used for finetuning the model $\mathbf{M}_{l, j_l}$. In this way, the model focuses on learning the representations for this subset of object classes. 

When learning the model $\mathbf{M}_{l, j_l}$, we use the model in its parent node as the initial point so that the knowledge from the parent node is transferred to the current model.
Since the root node is the pretrained for the 1000-class problem and finetuned for the 200+1 class problem, the model with larger level $l$ have inherited the knowledge from both 1000-class problem and the 200+1 class problem. 
Cascade is used for negative samples so that the model $\mathbf{M}_{l, j_l}$ focuses on hard examples that can not be handled well by the model $\mathbf{M}_{l-1, par(j_l)}$ in the parent node.

When finetuning the deep model, we use the multi-class cross-entropy loss to learn the feature representation. Then the 2-class hinge loss is used for learning the classifier based on the feature representation.

\section{Experimental results on the Hierarchical Feature Learning}

\subsection{Comparison with existing works}
We compare with single-model results across state-of-the-art methods. Table \ref{Table:allMethod} summarizes the result for RCNN \cite{girshick2014rich} and the results  from ILSVRC2014 object detection challenge. It includes the best results on the test data submitted to ILSVRC2014 from GoogLeNet, DeepID-Net, DeepInsignt, UvA-Euvision, and Berkeley Vision, NIN, SPP, which ranked top among all the teams participating in the challenge. Our model is based on the GoogLeNet model without adding any other layer and provided by the authors in \cite{ouyang2015deepid} online, which has mAP 40.3\%.\footnote{There are results higher than 40.3\% reported in \cite{ouyang2015deepid}, using additional layers, better region proposals, additional context and bounding box regression that we did not use but complementary to our approach. We use the 40.3\% baseline result they provide online to be consistent with the baseline introduced in Section \ref{sec:Baseline}.}
We also include the recent approach in \cite{yan2015object}. The approach in \cite{yan2015object} uses context (1.3\% mAP increase), better region proposal (0.9\% mAP increase in  \cite{yan2015object}), pair wise term for bounding box relationship (3.5\% mAP increase in  \cite{yan2015object}), which are not used  us but complementary to our all of implementations in Table \ref{Tab:Depth}.

\begin{table*}
\setlength{\tabcolsep}{1.5pt}
\centering
{\small
\begin{tabular}{*7cccc|c}
\hline	     approach  &  SPP$^*$   & NIN $^*$  & RCNN  &Berkeley &UvA 	& DeepInsight  & DeepID-Net &GoogLeNet & S-Pixels& ours \\
	     	        & \cite{he2014spatial} & \cite{lin2013network} & \cite{girshick2014rich}  &\cite{girshick2014rich} & \cite{Smeulders:SelectiveSearch}	&  \cite{Yan2014} & \cite{ouyang2015deepid}&\cite{szegedy2014going} & \cite{yan2015object} \\
%\hline   ImageNet val$_2$     &n/a & n/a &n/a&n/a & n/a&44.5  & ??\\
%\hline   ImageNet val$_2$ (avg)    &n/a & n/a &n/a&n/a & 42&45&44.5  & n/a\\
\hline   ImageNet val$_2$ &n/a &35.6  & 31.0 &33.4&n/a & 40.1&40.3&38.8 &44.8& 45.0\\
   ImageNet test      &31.8&n/a & 31.4  &34.5&35.4& 40.2& n/a &38.0 &42.5& n/a\\
\end{tabular}
}
\caption{Detection mAP ($\%$) on ILSVRC2014  for top ranked approaches with single model. For fair comparison with \cite{ouyang2015deepid}, we use their learned GoogLeNet parameters provided online as our baseline. The methods marked with * do not use classification data for pre-training.
%For ImageNet test, RCNN, Berkeley Vision and UvA-Euvision only reported single model result, while the remaining used model averaging in order to achieve the best results.   
}
\label{Table:allMethod}
\vspace{-5pt}
\end{table*}

\subsection{Ablation study}
The experiments in this section are only different in finetuining from the baseline introduced in Section \ref{sec:Baseline}.

\subsubsection{Investigation on different clustering methods}
\label{Sec:ExpClusterMethod}
%In this experiment, we empirically choose the one that has the best performance.
The experimental results that investigate different clustering methods are shown in Table \ref{Tab:Cluster}. In these experiments, we cluster the 200 object classes into 4 groups, which corresponds to the tree of level 2 in Fig. \ref{Fig:Cluster}. The models for the 4 groups are finetuned from the model finetuned using 200 object classes, which is the baseline with mAP 40.3\% introduced in Section \ref{sec:Baseline}. It can be seen from Table \ref{Tab:Cluster} that all the clustering approaches improved the detection accuracy except for the approach that randomly assigns the object classes into 4 groups, \emph{Random} in Table \ref{Tab:Cluster}. The use of confusion matrix and the use of visual similarity perform better than the other clustering approaches. When the wordnet id (WNID) is used, we cluster the 200 object classes into the following 4 groups:  1) animals and person; 2) device and traffic light; 3) instrumentation that is not device; 4) other remaining artifacts e.g. food, substance.

We find that the clustering results obtained from visual similarity is very similar to the results obtained from wordnet id for animals. We also find many examples of exceptions for artifacts and person. Person is assigned to artifacts that frequently contact with person, e.g. accordion, baby bed. Bookshelf, which belongs to instrumentation, is grouped with refrigerator, which does not belong to instrumentation in wordnet id. Baseball, which belongs to instrumentation, is grouped with bathing cap, which does not belong to instrumentation.

Experimental results show that the confusion matrix and the visual similarity have similar performance. We also find that their clustering results are very similar. Visually similar objects of different classes often cause confusion. Therefore, both confusion matrix and the visual similarity are good choices for clustering for our approach. Since the empirical results show that the visual similarity performs better than the other approaches, we have adopted it for clustering in our final implementation. 

\begin{table*}[]
\centering
{\small
\begin{tabular}{cccccccc}
\hline Clustering method & Random & Accuracy & Num Sample & Size   & WNID   & Confusion & Visual Sim \\
\hline
Increase in mAP       & -1.1\% & 0.48\%   & 0.56\%      & 0.66\% & 0.77\% & 0.91\%    & 0.967\%    
\end{tabular}}
\caption{Detection accuracy increase in mAP on ILSVRC val2 for different methods compared with the baseline model with mAP 40.3\%. The classification accuracy of object classes in the training data as the descriptor for clustering for \emph{accuracy}.  The number of samples is used as the descriptor for \emph{Num. Sample}. The average size of bounding boxes (measured by area of bounding box) is used as the descriptor for \emph{size}. The hierarchy in wordnet id (WNID) is used for clustering for \emph{WNID}. The confusion matrix of object classes is used as the similarity among classes for \emph{confusion}. The visual similarity in (\ref{eq:SoftMaxGrad}) is used as the similarity among classes for \emph{Visual Sim}. }
\label{Tab:Cluster}
\vspace{-10pt}
\end{table*}

\subsubsection{Investigation on the influence of hierarchy level}
The experimental results evaluating the influence of level $L$ in hierarchical feature learning  is shown in Table \ref{Tab:Depth}. Consistent mAP improvement is observed when the level increases. As the level increases from 1 to 4, the mAP increases from 40.3\% to 45\%. When the level increases, each model focuses on learning more specific feature representations. With the more specific representations, the features are more discriminative in distinguishing them from the background. Therefore, these better features lead to better detection accuracy.

Since we have adopted the bounding box rejection approach in \cite{ouyang2015deepid} for the model with level 1, there are only 136 boxes per image left for the model with level 1.
When the level is 4, there are 18 models to be evaluated for each bounding box. This seems to be a huge number. However, with the cascade, we can reject a large number of boxes for each model. On average, there are only 5.6 boxes per image evaluated for each model. Even if the 18 models are considered altogether, there are only around 100 boxes per image used for feature extraction and classification. Therefore, the use of multiple models that extract features for different object classes does not take much computational time.

\begin{table}[]
\centering
{\small
\begin{tabular}{c|cccc}
\hline Hierarchy level  $L$                           & 1      & 2      & 3      & 4     \\
\hline
\#. groups (=$N_m$)                       & 1      & 4      & 7      & 18    \\
avg \#. classes per group              & 200    & 50     & 29     & 11    \\
$N_{b,l}$      & 136    & 25.8   & 15.2   & 5.6   \\
$N_{b,l}\cdot N_m$     & 136    & 103.2  & 106.4  & 100.8 \\
mAP                               & 40.3\% & 41.3\% & 42.5\% & 45\% 
\end{tabular}}
\caption{Detection accuracy in mAP and other statistics on ILSVRC14 val2 for the cascaded hierarchical feature learning with different levels. $N_{b,l}$ denotes the average number of boxes per image evaluated per model for a given hierarchy level $l$.  $N_m$ denotes the number of models for a given tree depth.}
\label{Tab:Depth}
\end{table}

\begin{table}[]
\setlength{\tabcolsep}{2pt}
\centering
{\small
\begin{tabular}{c|c|c|c|c|c}
\hline
approach                             & $0 \!\Rightarrow \!1$ & $0 \!\Rightarrow \!2$      & $0\! \Rightarrow \!1 \Rightarrow \!2$      &  $0 \!\Rightarrow \!1 \Rightarrow\! 2 \!\Rightarrow \!3$       &  $0 \!\Rightarrow \!1 \!\Rightarrow \!3$   \\
\hline
mAP                                & 40.3\% & 38.9\% & 41.3\% & 42.5\% & 41.8\% 
\end{tabular}}
\caption{Detection accuracy in mAP on ILSVRC val2 for different finetuning strategies. 0 denotes the pretrained model, $l = 1, 2, 3$ denotes the model at the tree level $l$. For example,  $0 \Rightarrow 1$ denotes finetuning 200 object classes from the 1000-class model. $0 \Rightarrow 2$ denotes finetuning 4 class groups from the 1000-class model.  }
\label{Tab:Fintune}
\end{table}

\subsubsection{Investigation on the finetuning strategy}
In our final implementation, the deep models at higher levels (larger $l$) are finetuned based on the deep model at lower levels. The experimental results in Table \ref{Tab:Fintune} shows the variations on the finetuning strategy. 

If we use only one model for 200 classes and finetune this model from the 1000-class ImageNet pretrained model, the performance is 
40.3\%, which the baseline described in Section \ref{sec:Baseline} and denoted by  $0 \Rightarrow 1$ in Table \ref{Tab:Fintune}. 

If we directly finetune the 4 models at level 2 from the 1000-class pretrained model, the mAP decreases to 38.9\%.  In comparison, when learning 4 models by using the 200-class finetuned model at level 1 as the initial point, the mAP increases to  41.3\%.  The 4 models at level 2 focuses on discriminating around 50 object classes from the background. 
Direct finetunng of the 4 models from the pretrained model does not use the knowledge of their correlated other 150 object classes. In comparison, finetuning from the model at level 1, which is finetuned using the 200 object classes, has used the knowledge from the 200 object classes. Therefore, improvement is observed when using the 200-class finetuned model as initial point. 

When finetuning the models at level 3, the use of the models at level 2 as initial point has mAP 42.5\%, and the use of the models at level 1 as initial point has mAP 41.8\%. The learning strategy that gradually focuses the model from 200 to 50 and then to 29 classes performs better than the abrupt jump from 200 classes to 29 classes. 

\subsubsection{Results on the PASCAL VOC}
We also observe 1.2\% mAP improvement on PASCAL VOC 2007 when its object classes are clustered into 4 groups for GoogLeNet.

\section{Conclusion}
This paper provides analysis and experimental results on the factors that influences finetuning on the object detection task. We find that it is better to have the number of samples uniform across different classes for feature learning. A cascaded hierarchical feature learning is proposed to improve the effectiveness of the learned features. 4.7\% absolute mAP improvement is achieved using the proposed scheme without much increase in computational cost.

\textbf{Acknowledgment:} This work is supported by SenseTime Group Limited and the General Research Fund sponsored by the Research Grants Council of Hong Kong (Project Nos. CUHK14206114, CUHK14205615, CUHK417011, and CUHK14207814).

{\small
\bibliographystyle{ieee}
\bibliography{./PME}
}

\end{document}